\documentclass[runningheads]{llncs}
\usepackage{eccv}
\usepackage{eccvabbrv}
\usepackage{graphicx}
\usepackage{booktabs}
\usepackage{arydshln}
\usepackage{svg}
\usepackage{amsmath} 
\usepackage{algorithm}
\usepackage{algpseudocode}
\usepackage[accsupp]{axessibility}

\newcommand{\ignore}[1]{}

\usepackage{hyperref}

\usepackage{orcidlink}

\begin{document}

\title{Enhanced Automotive Object Detection via RGB-D Fusion in a DiffusionDet Framework} 

\titlerunning{RGB-D FusedDiffusionDet}

\author{Eliraz Orfaig\inst{1}\orcidlink{0000-1111-2222-3333} \and
Inna Stainvas\inst{2,3}\orcidlink{1111-2222-3333-4444} \and
Igal Bilik\inst{3}\orcidlink{2222--3333-4444-5555}}

\authorrunning{Orfaig et al.}

\institute{School of Electrical and Computer Engineering, Ben Gurion University of the Negev, Beer Sheva, Israel.}

\author{Eliraz Orfaig, {\it Student Member, IEEE}, Inna Stainvas and
Igal Bilik, {\it Senior Member, IEEE}\thanks{Eliraz Orfaig and Igal Bilik are with the School of Electrical and Computer Engineering, Ben Gurion University of the Negev, Beer Sheva, Israel. (e-mails: elirazo@post.bgu.ac.il, bilik@bgu.ac.il). Inna Stainvas is a Senior Scientist - AI, GE Healthcare Science \& Technology, (e-mails: Inna.stainvas@gehealthcare.com)}
}
\maketitle

\begin{abstract}
  Vision-based autonomous driving requires reliable and efficient object detection. This work proposes a DiffusionDet-based framework that exploits data fusion from the monocular camera and depth sensor to provide the RGB and depth (RGB-D) data. Within this framework, ground truth bounding boxes are randomly reshaped as part of the training phase, allowing the model to learn the reverse diffusion process of noise addition. The system methodically enhances a randomly generated set of boxes at the inference stage, guiding them toward accurate final detections. By integrating the textural and color features from RGB images with the spatial depth information from the LiDAR sensors, the proposed framework employs a feature fusion that substantially enhances object detection of automotive targets. The $2.3$ AP gain in detecting automotive targets is achieved through comprehensive experiments using the KITTI dataset. Specifically, the improved performance of the proposed approach in detecting small objects is demonstrated.

\end{abstract}

\section{Introduction} \label{sec:intro}
The global advanced driver-assistance systems (ADAS) and the autonomous driving market have grown exponentially for the last decade~\cite{6586127}. ADAS is expected to reduce traffic accidents and casualties dramatically~\cite{8633345,7317855}. Autonomous driving systems require excellent detection performance in a broad spectrum of automotive scenarios~\cite{9127857, 9760734}.   
The success of the ADAS and autonomous driving transformation largely depends on two components: (i) sensing suit that acquires information on autonomous vehicles' surroundings~\cite {6586127, 8378587,6586127, 7485215, 10371070} and (ii) deep learning to process the collected measurements~\cite{8828025,10438394, 8835792}. 

Autonomous vehicles decide on their actions according to the information on their surroundings provided by the sensing suit. The conventional autonomous vehicle sensing suit consists of light detection and ranging (LiDARs)~\cite{9760734}, radars~\cite{6586127}, and stereo or mono  cameras~\cite{8828025}. RGB images captured by monocular or stereo cameras provide high spatial resolution, color, and texture on the surrounding objects. However, the large variability of object appearances challenges vision-based decision-making~\cite{9270494, 8661069}. This challenge can be addressed by fusing the RGB data with depth measurements provided by the LiDAR or radar sensors. However, depth sensors have limited spatial resolution and dynamic range~\cite{8944496, 10324930, 9251080}. Therefore, the fusion of RGB images with depth information (RGB-D) is needed to enhance the object detection robustness and accuracy in complex urban environments~\cite{ZHANG2023146}, where the accurate identification and localization of a wide range of objects, such as pedestrians, vehicles, and obstacles, are crucial for safe navigation.

\begin{figure}[tb]
    \hspace*{0.7cm} 
    \includegraphics[width=10cm]{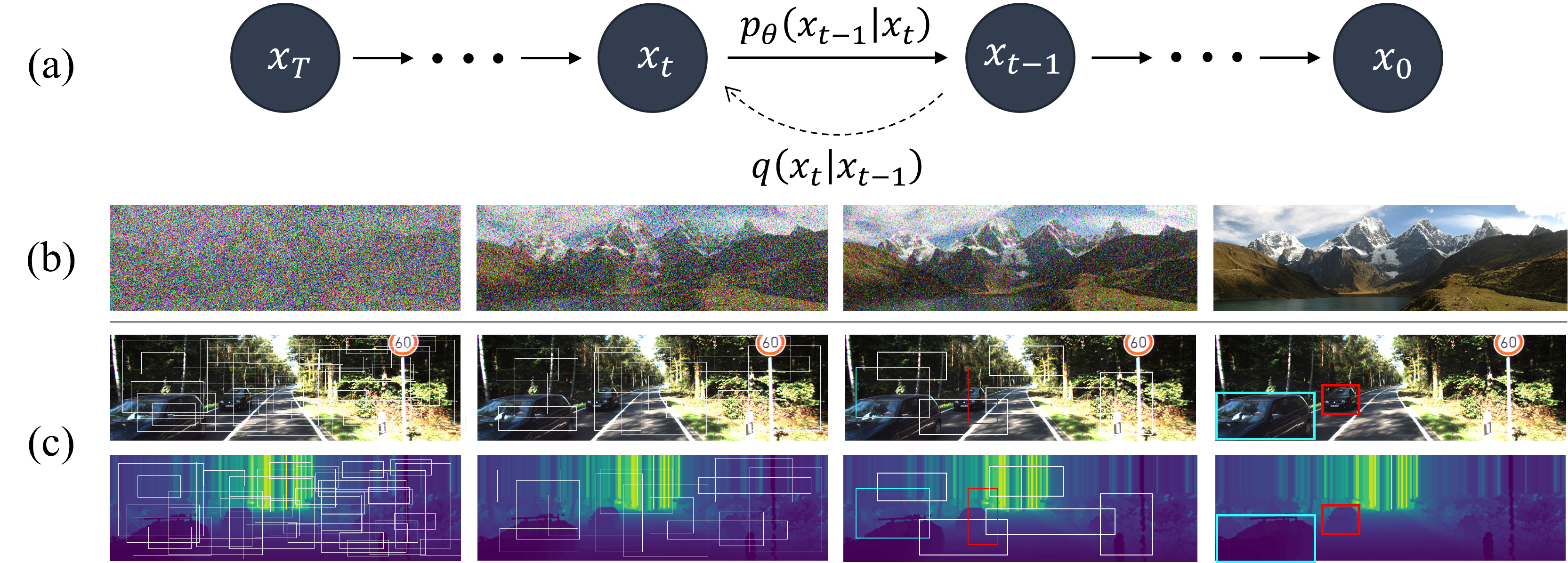}
    \caption{An RGB-D diffusion model for object detection. (a) A diffusion model where $q$ is the diffusion process and $p_\theta$ is the learnable reverse process. (b) Diffusion model for image generation task. (c) Object detection is a denoising diffusion process from noisy boxes to object boxes on the RGB-D dataset.}
    \label{fig1}
\end{figure}

Recently introduced deep neural networks (DNN) have dramatically enhanced monocular vision object detection capabilities \cite{WU202039, SparseRCNN, girshick2014regions ,FastRCNN ,FasterRCNN, YOLOv7}, which was further improved by transformer architectures \cite{DETR,FastConDETR,DNDETR,SparseRCNN,DeformableDETR}. 
The most recently proposed DiffusionDet model~\cite{chen2023diffusiondet} is based on diffusion process theory. This model was originally introduced for RGB data only. Our work generalizes the DiffusionDet model for multi-modal RGB-D data processing. The main idea of the proposed approach is to introduce noise-to-box diffusion processes~\cite{chen2023diffusiondet}, initiated with purely random boxes, and gradually refine their positions and sizes until they perfectly cover the targeted objects. Fig.~\ref{fig1} shows the exemplary noise-to-image diffusion process~\cite{ho2020denoising}, implemented in analogy with the noise-to-box diffusion model-based denoising~\cite{dhariwal2021diffusion, ho2020denoising, song2020score}. These models can be seen as a class of likelihood-based models for image generation, which gradually remove image noise via the learned denoising process.

This work proposes the FusedDiffusionDet architecture that generalizes the DiffusionDet for RGB-D data processing. The proposed architecture achieves statistically significant object detection performance improvement of $3.7\%$ for small objects, such as pedestrians, and $2.9\%$ for large objects, such as vans.  
The proposed FusedDiffusionDet architecture can be generalized straightforwardly to any $2$D data acquired by various sensors, such as event or thermal cameras or additional optical sensors. In addition, the proposed approach considers the pre-trained architectures, which simplify and accelerate training procedures and provide higher and more reliable object detection performance.
The main novelty of the proposed approach can be summarized as follows:
\begin{itemize}
    \item FusedDiffusionDet architecture that generalizes the DiffusionDet for object detection using multi-modal data.
    \item Ablation studies on various feature fusion architectures for object detection.
    \item Efficient training of FusedDiffusionDet network using refinement only. 
\end{itemize}

The rest of this article is organized as follows. Section~\ref{RW} summarises the related work. Section~\ref{Sec:PA} presents the proposed diffusion network-based object detection approach. Section~\ref{Exp} details the conducted experiments, and Section~\ref{Res} evaluates the object detection performance of the proposed approach. Our conclusions are summarised in Section~\ref{Concl}.

\section{Related work}\label{RW}
Conventionally, autonomous vehicles fuse information from multiple sensing modalities to achieve the required robustness. Three types of sensor fusion approaches that consider the fusion of sensed information at various processing stages were studied in the literature~\cite{Boulahia2021EarlyIA}. Early fusion approaches consider the fusion of the raw data from various sensors, providing a comprehensive dataset for preliminary analysis~\cite{Haris2022NavigatingAA, Lim2019RadarAC}. Intermediate fusion approaches, also denoted as feature-level fusion, consider the fusion of extracted features from individual sensors to create a more explicit object representation~\cite{Wang2023VIMIVM}. Late fusion considers the object-level fusion of independently processed data at each sensor modality to leverage their unique capabilities~\cite{Ding2022AMF}. Recently, the hybrid approaches that combine sensor fusion at various processing stages and leverage their advantages were proposed in the literature~\cite{ShahianJahromi2019HybridSF}.

DNNs became the primary approach for monocular vision processing~\cite{WU202039, SparseRCNN, girshick2014regions ,FastRCNN ,FasterRCNN, YOLOv7}. However, poor lighting and adverse weather conditions, diversity of road infrastructures and vulnerable user appearances, and real-time processing requirements challenge the practical implementation of state-of-the-art DNN-based processing~\cite{ZHANG2023146}. 
These challenges can be addressed by the DNN-based fusion of RGB images with data from other sensor modalities~\cite{S2022DeepLN,khan2018multimodal,lee2019mv3d}. Similarly to its success in image processing, DNN-based sensor fusion was shown to provide robust and accurate object detection in various environmental conditions.
Over the last decade, multiple DNN-based object detection approaches for automotive applications were studied in the literature~\cite{s23104832}. However, a large variety of objects in typical automotive images challenges their recognition. Object detection in optical images depends on their shape, size, color, and context within the image. 
The conventional DNN-based object detection approaches rely on prior information about the desired objects. They are conventionally formulated as a regression or classification on empirically designed object candidates, using manually structured techniques such as sliding windows~\cite{GirshickDDM13,Sermanet2013OverFeatIR}, region proposals~\cite{FastRCNN,FasterRCNN}, anchor boxes~\cite{FocalLoss,YOLOv7}, and reference points~\cite{Duan_2019_ICCV,RepPoints, ObjectsAsPoints}. 

Recently, the learnable object queries were proposed for object detection~\cite{SparseRCNN, DeformableDETR}. The detection transformer (DETR) architecture has introduced an end-to-end detection pipeline using a learnable object queries paradigm, which does not consider manually-designed object features~\cite{DETR,FastConDETR,DNDETR,SparseRCNN,DeformableDETR}. The major drawback of this paradigm is the requirement to define a fixed set of learnable queries. 

Diffusion models, inspired by nonequilibrium thermodynamics, have gained significant attention for their ability to capture and simulate complex processes \cite{ho2020diffusion}. The diffusion approach systematically analyzes observed data by introducing noise and subsequently reconstructs the original data by preserving this iterative process. Recently, the diffusion approach was adopted for various applications~\cite{Nichol2021Glide, Ho2021Scaling, Ramesh2023Imagen,Bai2023DALLE3}. The major challenge in generative modeling lies in balancing their tractability and flexibility. Tractable models can be analytically evaluated and inexpensively fit data, but their ability to describe the structure in rich datasets may be limited. On the other hand, flexible models can fit arbitrary data structures but are computationally complex for evaluation, training, and sampling. The major benefit of the diffusion models is that they achieve both analytical tractability and flexibility~\cite{nichol2021improved, ho2020denoising}.

\section{The Proposed Approach}\label{Sec:PA}
This section summarises the proposed diffusion network-based sensor fusion approach for object detection. The proposed approach enables the fusion of the RGB images from the monocular camera with any auxiliary depth sensor whose data can be represented as $2$D images. This work uses the KITTI dataset~\cite{Geiger2012CVPR}, where depth information is provided by the Velodyne HDL-64E LiDAR. The LiDAR $3$D points are first pre-processed to align with the RGB images from the camera and then converted to dense  $2$D depth images. 

\subsection{Architecture}
\subsubsection{Framework:}
\begin{figure}[t]
    \includegraphics[width=12cm]{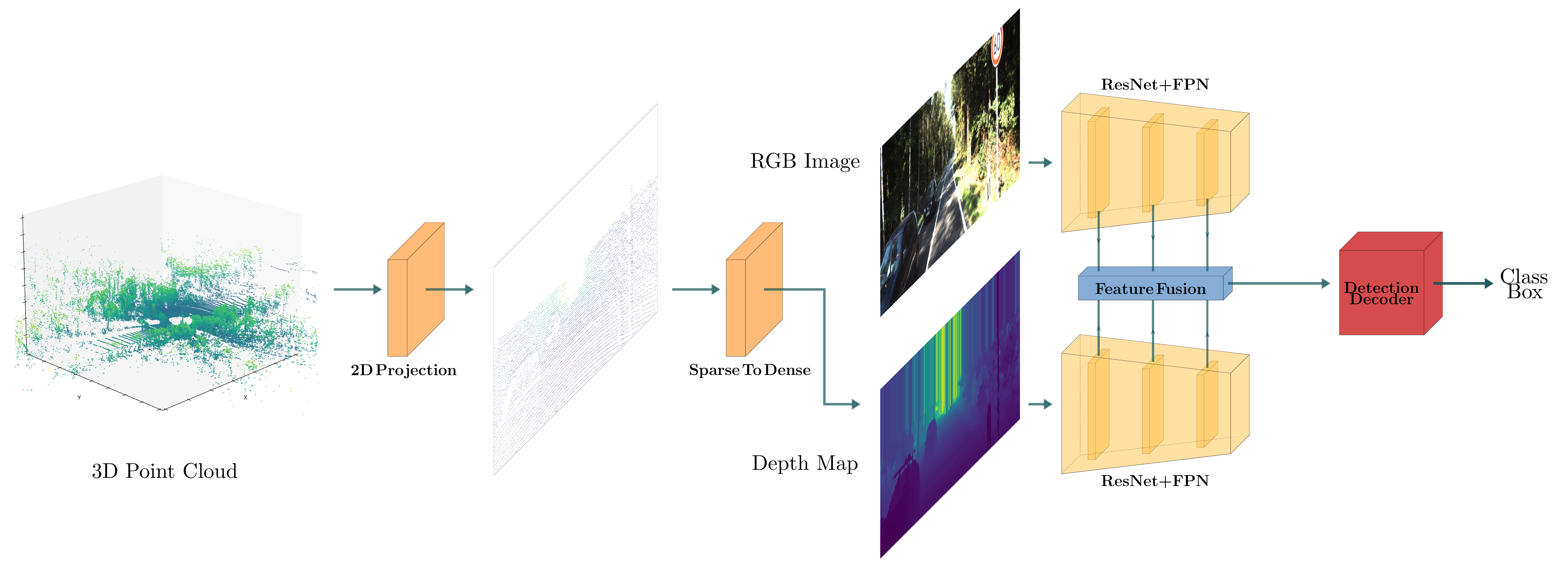}
    \caption{Schematic representation of the proposed diffusion-based object detection architecture using RGB-D data. A $3$D point cloud is transformed into a $2$D image, enriching it with depth information. The diffusion detection model~\cite{chen2023diffusiondet} network uses fused features to detect and classify objects. The output is a set of bounding boxes that indicate the location and classification of objects within the scene.}
    \label{fig2}
\end{figure}

Fig.~\ref{fig2} shows the proposed network architecture for object detection using the fused information from the RGB images and LiDAR point clouds in the diffusion framework. 
This architecture enables multi-scale feature extraction and is motivated by its ability to capture both high-level semantics and fine-granularity details.
The proposed architecture contains two processing branches for the monocular camera and the LiDAR that encodes the RGB images and depth information into feature space. 
The RGB images contain the object's visual appearance, texture details, and color information needed for identifying the visual aspects of the scene. The LiDAR point cloud processing provides depth maps with spatial and geometric information on the object's shape and location.   
The proposed architecture contains two separate ResNet-50~\cite{he2016deep} backbones, augmented with individual feature pyramid networks (FPNs)~\cite{lin2017feature}, fitted for each processing pipeline. 
The network architecture in Fig.~\ref{fig2} contains the feature fusion that integrates the outputs of the two FPNs and, thus, combines the advantages of RGB and depth information, resulting in a rich, unified feature representation. The fused data is processed in the detection head for object detection and classification.

\subsubsection{Preprocessing:} 
LiDAR emits laser beams and captures the reflections from the surrounding objects. The reflections are represented as a $3$D point cloud in the LiDAR's coordinates and the reflection intensity. A monocular camera outputs $2$D images by projecting the $3$D environment onto a camera plane in the camera's coordinates. 
Therefore, LiDAR and the camera capture their surroundings using different coordinate systems, and data preprocessing, including temporal synchronization and spatial alignment, is needed for their data fusion. The considered KITTI dataset~\cite{Geiger2012CVPR} provides calibration parameters needed for this transformation. It involves calculating the necessary rotation and translation to accurately project the LiDAR's $3$D point cloud onto the camera's $2$D image plane. The outcome of this projection is a sparse depth map with multiple discrete points with depth information. 
The classical low-cost RGB guided depth densification method~\cite{ku2018defense} can be used to complete the sparse depth image. 
\ignore{In the proposed approach, DNN is used to convert sparse depth images to dense ones by completing the missing depth information. The depth completion process, from~\cite{ku2018defense}, employs classical, low-cost complexity image processing approaches such as dilation, extrapolation, and hole-filling to achieve this.}

\subsubsection{Diffusion Model for Object Detection:}Diffusion models are inspired by nonequilibrium thermodynamics principles~\cite{sohl2015deep} and form a Markovian chain through a forward diffusion process that systematically introduces noise into data samples:
\begin{equation}
q(x_t|x_0) = \mathcal{N}(x_t | \sqrt{\bar{\alpha}_t} x_0, (1 - \bar{\alpha}_t) \mathbf{I})\;,\label{Eq1}
\end{equation}
where $x_0$ denotes the initial bounding boxes, $x_t$ represents the  noisy bounding boxes at the time instance, $t$, and $\bar{\alpha}_t$ is the product of noise levels over time. Notice that the data in~\eqref{Eq1} is incrementally shifted and reshaped via the Markov chain mechanism towards a Gaussian distribution, emphasizing transformation from precise ground truth bounding boxes to their noisy equivalents.

During the training, a neural network, \( f_{\theta}(x_t, t | z_x) \), learns the reverse of this diffusion process. The objective is to accurately predict the original bounding box, \( x_0 \), from its noisy version \( x_t \), given the input extracted features, $z_x$, inside bounding boxes, $x$. The training is performed by optimizing the following loss function:
\begin{equation}
\mathcal{L}_{\text{train}} = \frac{1}{2} \|f_{\theta}(x_t, t | z_x) - x_0\|^2\;,
\end{equation}
which guides the network in minimizing the difference between the reconstructed and ground truth data and, in this way, enhances the model's ability to generate high-fidelity data samples.

During the inference, the trained neural network iteratively reconstructs the original data from its noisy counterpart, demonstrating the model's efficacy in capturing and generating complex data distributions.

\subsubsection{Feature Fusion:}\label{SS:FF}
\begin{figure} [tb]
    \centering
    \hspace*{-2.5cm} 
    \includegraphics[width=14cm]{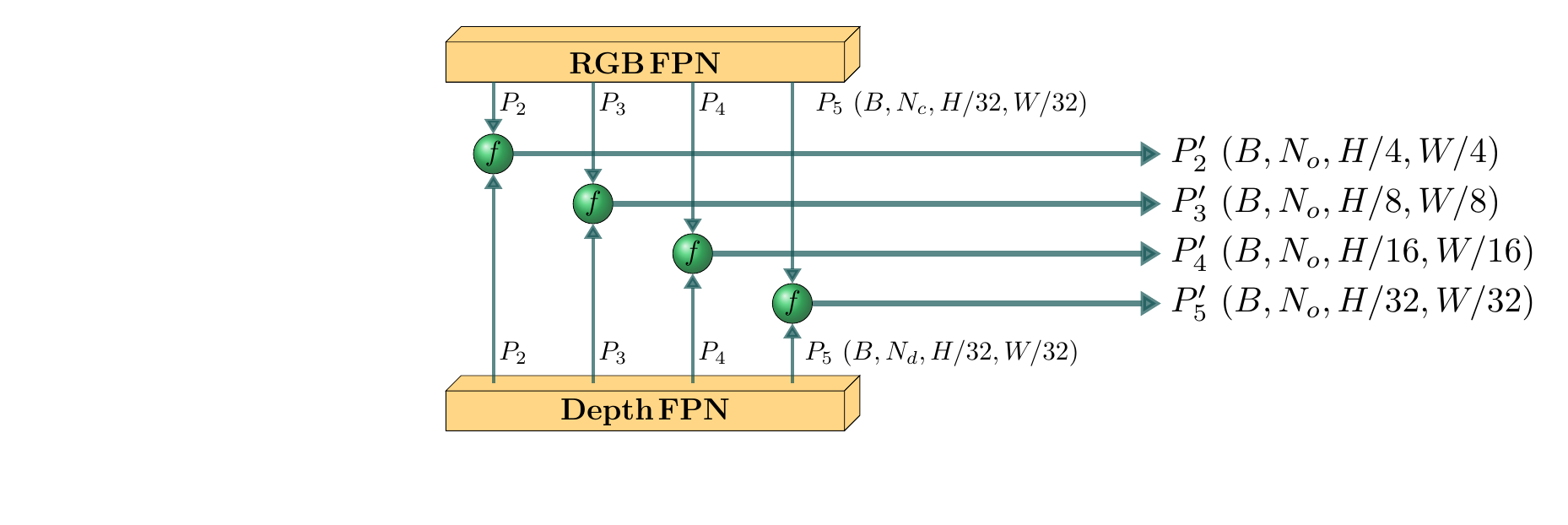}
    \caption{Feature fusion processing, where the green nodes denote fusion operation.}
    \label{fig3}
\end{figure}

Feature fusion is a key component of the proposed approach that leverages the complementary information from RGB and depth data for object detection. 
The outputs of $P_2-P_5$ FPNs that contain $256$ channels are fused at this stage, as shown in Fig.~\ref{fig3}. 
Various fusion operation types, concatenation ($\oplus$), small convolution net ($conv$), summation ($sum$), spatial multi-layer perceptron ($MLP$) net, and cross-attention mechanisms ($ca$), are considered in this work. 

\ignore{The concatenation of the corresponding FPN outputs from the RGB and depth processing pipelines provided an optimal and computationally efficient approach. The superiority of concatenation fusion can be explained by its ability to avoid any possible information loss in other fusion approaches. According to this fusion approach, the $P2, P3, P4$, and $P5$ from the RGB processing are concatenated with their corresponding FPNs from the depth processing, resulting in new tensors for each pair of $P2, P3, P4$, and $P5$. Each fused tensor contains $512$ channels, thus preserving the data's spatial dimensionality and doubling its channel capacity for enhanced data representation.}

\subsubsection{Fusion Network Architectures:}
The cross-attention mechanism as an architecture and two fusion network architectures, dubbed $MLP$ in Algorithm~\ref{A:DS} and $conv$ in Algorithm~\ref{A:SPS} are considered in this work. The conventional cross-attention architecture with RGB tensor $x_c$ as a query ($Q$), depth $x_d$ as the key ($K$), and value ($V$) as described in Algorithm~\ref{A:CA} is used in this work.

The input to both fusion network architectures is the fused concatenated RGB-D feature, $x_{\oplus}=\oplus(x_{c}, x_{d})$, where $x_c$ and $x_d$ are RGB and depth features, respectively. Assume shapes of  $x_{c}$ and $x_d$ being $[B, N_{c}, H, W]$ and $[B, N_{d}, H, W]$, respectively; the shape of a concatenated tensor is $shape(x_{\oplus})=[B, N=N_c+N_d, H, W]$. The fusion operation goal is to reduce the dimensionality of the fused tensor to some desired dimensionality, $N_o$.
Note that $conv2d$ operation in Algorithms~\ref{A:DS}-\ref{A:SPS} is similar to PyTorch $Conv2d(N_{in}, N_{out},k,s=1)$, where $N_{in}$ is a number of channels in the input tensor, $N_{out}$ is the number of channels in the output tensor, $k$ is the kernel size, $s$ is the stride parameter, with the default value $1$, when not indicated. The lattice size $H,\; W$ remains constant. Therefore, the appropriate padding is performed where required. 

\begin{algorithm}[tb]
\caption{Feature Fusion: Cross-Attention}
\label{A:CA}
\begin{algorithmic}[1]
\Require Reshape $x_{c}$ and $x_{d}$ to $[HW, B, N_{c}]$ and $[HW, B, N_{d}]$
\Statex \textbf{Processing Layers:}
\State $x_{attn} \gets \text{Attention}(Q=x_{c}, K=x_{d}, V=x_{d})$
\State $x_{residual} \gets x_{c} + x_{attn}$
\State $x_{out} \gets \text{LayerNorm}(x_{residual})$
\State Reshape $x_{out}$ to $[B, N_{o}, H, W]$
\Ensure Output tensor $x_{out}$ of shape $[B, N_o, H, W]$
\end{algorithmic}
\end{algorithm}

\begin{algorithm}
\caption{Feature Fusion: $MLP$}
\label{A:DS}
\begin{algorithmic}[1]
\Require concatenated tensor $x_{\oplus}$
\Statex \textbf{Processing Layers:}
\State $x_{1} \gets (ReLU \circ conv2d(N, N, k=1))[x_{\oplus}]$
\State $x_{2} \gets (ReLU \circ conv2d(N, N_o, k=1))[x_1]$ 
\State $x_{out} \gets conv2d(N_o, N_o, k=1, s=1)[x_2]$ 
\Ensure Output tensor $x_{out}$ of shape $[B, N_o, H, W]$
\end{algorithmic}
\end{algorithm}

\vspace{-1cm}

\begin{algorithm}
\caption{Feature Fusion: $conv$}
\label{A:SPS}
\begin{algorithmic}[1]
\Require concatenated tensor $x_{\oplus}$ 
\Statex \textbf{Processing Layers:}
\State $x_1 \gets (ReLU \circ conv2d(N, N, k=3))[x_{\oplus}]$
\State $x_2 \gets conv2d(N, N_o, k=3)[x_1]$
\State $x_3 \gets conv2d(N, N_o, k=1) [x_{\oplus}]$
\State $x_{out} \gets x_{2} + x_{3}$
\Ensure Output tensor $x_{out}$ of shape $[B, N_o, H, W]$
\end{algorithmic}
\end{algorithm}

\subsection{Loss Function}
This subsection details the considered here loss function. The total loss function, $L_{\text{Total}}$ is defined as a weighted sum of classification, $\lambda_1 L_{\text{cls}}$ and regression, $\lambda_2 L_{\text{reg}}$ losses:
\begin{equation}
L_{\text{Total}} = \lambda_1 L_{\text{cls}} + \lambda_2 L_{\text{reg}}\;,
\end{equation}
where $\lambda_1$ and $\lambda_2$ are the weights of the classification and regression losses, respectively. The classification loss $L_{\text{cls}}$ is computed using the focal loss function~\cite{lin2017focal}, which is designed to address the class imbalance problem by modulating the loss contribution from each sample:
\begin{equation}
L_{\text{cls}} = -\sum_i \left[ v_i (1 - \hat{v}_i)^\gamma \log(\hat{v}_i) + (1 - v_i) \hat{v}_i^\gamma \log(1 - \hat{v}_i) \right]\;,
\end{equation}
where $v_i$ represents the ground truth class probability, $\hat{v}_i$ is the predicted class probability, and $\gamma$ is a modulating factor that controls the weight given to each example.

The regression loss $L_{\text{reg}}$ is a combination of $L1$ Loss, generalized intersection over union (IoU) loss~\cite{rezatofighi2019generalized}, and center loss, each contributing to different aspects of bounding box prediction accuracy:
\begin{equation}
L_{\text{reg}} = \lambda_3 L_{L1} + \lambda_4 L_{\text{GIoU}}\;,
\end{equation}
where $\lambda_3$, $\lambda_4$ are weights assigned to $L1$ Loss, and generalized IoU loss, respectively. This combination of losses ensures accurate prediction of the bounding box location, size, and orientation, leading to more precise object detection.

\section{Experiments}\label{Exp}
This section presents the experimental design and the evaluation metrics to assess the experiments' performance. 
The performance of the proposed approach with DiffusionDet architecture and data fused from RGB and depth sensing modalities is evaluated in Section~\ref{Res}. The performance of the proposed approach is also compared with the DiffusionDet detection performance used for each sensing modality independently. 

\subsection{Experimental Design}\label{SS:ED}
The advantages of the object detection performance using the fused RGB-D data are emphasized in this work by comparing the performance of the fused RGB-D network with the unimodal RGB and depth processing networks, denoted as ${\cal N}_{t}^{n}$, where the input type, $t\in\{c,\;d\}$ can be $c$ for RGB data or $d$ for depth $n$ is the number of output feature channels in $P_i$ of FPN's blocks. The fused model is denoted as ${\cal N}_{f(c_{n_c}, d_{n_d})}^{n}$, where $f$ is the fusion operation, $n_c,~n_d$ are the numbers of feature tensor channels in parallel color and depth $P_i$ blocks and $n$ is the number of channels of the fused feature tensor.

The fusion operations such as concatenation ($\oplus$), convolution network ($conv$), summation ($sum$), spatial multi-layer perceptron network ($MLP$), and cross-attention ($ca$)  were considered in this work (see Section~\ref{SS:FF} for details). 
Each operator provides a different strategy for integrating the features acquired by the RGB and depth sensors. 
In this work, the unimodal RGB DiffusionDet, ${\cal N}_{c}^{256}$, with $256$ output channels, and the RGB-D FusedDiffusionDet, ${\cal N}_{\oplus(c_{256},d_{256})}^{512}$, with $256$ features channels in $P_i$, and $512$ output feature channels were implemented. Notice that the super-script index reflects the network capacity.

\subsection{Dataset}
The KITTI dataset \cite{Geiger2012CVPR}, designed especially for autonomous driving, provides high-resolution RGB images, raw LiDAR scans, and precise calibration data, facilitating a multi-modal approach to sensor integration. It contains $7.4K$ annotated images with a variety of object classes, such as vehicles, pedestrians, and cyclists.
The training dataset is divided into disjoint training and testing sets, where $85\%$ of the images are used for training and $15\%$ for performance evaluation.\footnote{Notice that no overfitting phenomenon was observed in our experiments, and the network training was stopped when the training loss plateau was reached. This behavior can be explained by training initializing models from pre-trained backbones, which is expected to reduce network capacity implicitly.}

\subsection{Implementation Details} 
The ResNet backbones are used for both RGB and depth feature extraction. Since our depth encoder uses the ResNet backbone, it is crucial to transform a single depth image to an RGB one\footnote{Our unpublished experiments demonstrated that using the single depth channel results in degraded performance.}. We use the viridis colormap of the Matplotlib library, commonly applied in depth-related works to improve depth perception. The ResNet backbones for RGB and depth processing were initialized using pre-trained weights from ImageNet-1K~\cite{deng2009imagenet}. The model was trained for over $60K$ iterations on an NVIDIA RTX 4090 GPU, opting for a mini-batch size of $2$. Data augmentation is restricted to random horizontal flips, and the number of proposal boxes was set to $N_{train}=500$.

\subsection{Performance Evaluation Metrics}
Conventionally, the object detection approaches use the average precision (AP) metric, which comprehensively assesses accuracy across various levels of precision and recall~\cite{henderson2017endtoend}. AP accounts for the mean precision value for different recall thresholds, providing a singular performance measure. AP$_{50}$ and AP$_{75}$ metrics were calculated to assess localization accuracy at the IoU thresholds of $0.5$ and $0.75$, respectively. AP$_s$, AP$_m$, and AP$_l$ are size-specific metrics for small, medium, and large objects, addressing the model's capability to detect objects across various sizes.
These variants enable a nuanced evaluation of model performance, considering both the precision in object localization and the effectiveness in detecting objects of varying sizes.

\section{Results and Analysis}\label{Res}
\begin{figure}[tb]
    \centering
    \includegraphics[width=0.8\textwidth]{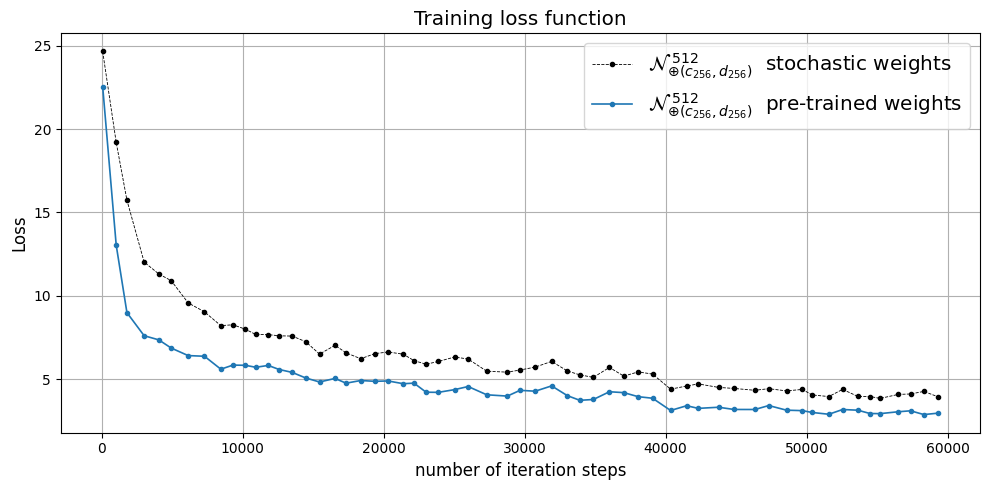}
    \caption{Comparison of training loss functions with random and pre-trained backbones. A pre-trained backbone significantly enhances performance and reduces computational time compared to the initialization of random weights. Therefore, all our experiments use pre-trained backbones.}
    \label{fig6}
\end{figure}

The performance of the proposed approach is evaluated in this section. The influence of data type, model capacity, and fusion strategy on the detection performance of the proposed networks is evaluated in the following subsections.
Figure~\ref{fig6} shows that training with pre-trained backbones outperforms that with stochastic weight initialization. Therefore, all evaluated models start training from a pre-trained backbone. Additionally, in fused models, both RGB and depth encoder backbone networks are initialized with identical pre-trained weights to ensure consistency in feature extraction for different sensor modalities.

This section is organized as follows. First, the unimodal models' performance is evaluated in Subsection~\ref{Uni}. Next, Subsection~\ref{LLF} describes low-level fusion experiments. Various considered fusion architectures are evaluated in Subsection~\ref{insight}. Finally, the best fused-model performance is shown in Subsection~\ref{opl}.

\subsection{Unimodal Models Performance}\label{Uni}
The depth and RGB models of different capacities are presented in the first three lines of Table~\ref{tab:object_detection_performance}.
Comparison of unimodal models of the same capacity of $N_{c}^{256}$ and $N_{d}^{256}$ shows that RGB data is more valuable than depth data for detection of all tested object categories. The RGB model $N_{c}^{256}$ outperforms $N_{d}^{256}$ in all performed experiments, with the most significant gap observed for small objects with the $AP_s$ degradation from $60.5\%$ to $36.7\%$. 
The depth sensor's lower resolution is one reason for this performance gap.

Comparing the RGB models $N_{c}^{256}$ and $N_{c}^{512}$ in lines $2$ and $3$ in Table~\ref{tab:object_detection_performance}, notice that the increased model's capacity results in only insignificant performance improvement. Expanding the number of RGB features from $256$ to $512$ channels has slightly improved the AP from $56.1\%$ to $56.9\%$.
This insignificant improvement indicates that the RGB encoder has almost reached the required capacity for solving the considered detection tasks, and additional RGB features are inefficient.

\subsection{Low Level Fusion}\label{LLF}
\begin{figure}[tb]
    \centering
    \includegraphics[width=0.8\textwidth]{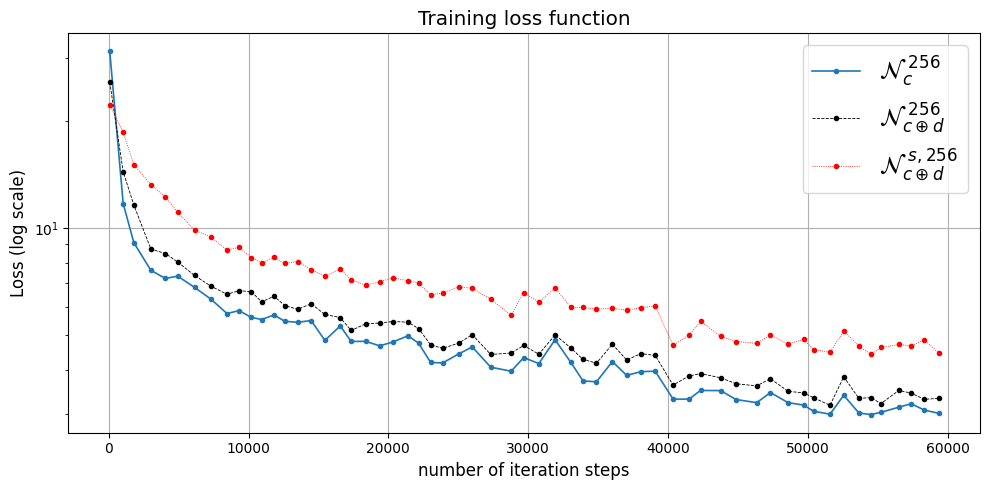}
    \caption{Training loss functions for RGB and low-level fused RGB-D models. Low-level fusion does not improve detection performance. Notice that ${\cal N}_{c \oplus d}^{256}$ with a partial usage of a pre-trained backbone model outperforms ${\cal N}_{c \oplus d}^{s,256}$.}
    \label{fig5}
\end{figure}

This work considered the network architecture with ResNet backbone and RGB-D data concatenated in the input. This work tested two training strategies: (i) starting from stochastic weights in all layers and (ii) starting from stochastic weights in the first layer and weights from pre-trained ResNet in all the rest layers. The models obtained from these training strategies are denoted as  ${\cal N}_{c \oplus d}^{s,256}$ and ${\cal N}^{256}_{c \oplus d}$, respectively. Fig.~\ref{fig5} shows that direct concatenation with depth does not improve object detection performance. This observation may be explained by the errors in the depth channel due to converting sparse depth into a dense one. Notice that even partial usage of the pre-trained backbone model is advantageous, leading to better performance of model ${\cal N}_{c \oplus d}^{256}$ compared with ${\cal N}_{c \oplus d}^{s,256}$.      

\begin{table}[tb]
    \caption{Object detection performance of models using AP metrics at different IoU thresholds ($50$ and $75$) and object sizes ($s$, $m$, and $l$). The first three lines stand for unimodal models, and the fourth is for the best-fused model.}
    \label{tab:object_detection_performance}
    \centering
    \begin{tabular}{@{}lcccccc@{}}
    \toprule
    Model & ~~~AP & AP$_{50}$ & AP$_{75}$ & ~~~AP$_s$ & AP$_m$ & AP$_l$ \\ 
    \midrule
    ${\cal N}_d^{256}$ & ~~~44.6 & 72.5 & 45.9 & ~~~36.7 & 44.2 & 50.7 \\
    ${\cal N}_c^{256}$ & ~~~56.1 & 81.1 & 61.7 & ~~~60.2 & 54.9 & 61.2 \\
    ${\cal N}_c^{512}$ & ~~~56.9 & 82.4 & 61.1 & ~~~60.4 & 55.6 & 62.2 \\
    ${\cal N}_{\oplus(c_{256},d_{256})}^{512}$  & ~~~\textbf{58.7} & \textbf{83.2} & \textbf{65.9} & ~~~\textbf{60.5} & \textbf{58.1} & \textbf{62.9} \\
    \bottomrule
    \end{tabular}
\end{table}

\begin{table}[tb]
    \caption{The object detection performance of evaluated mid-level fusion approaches using AP metric. Original and fused encoder feature sizes are identical and equal to $256$. The blue color indicates the performance of the original RGB unimodal, ${\cal N}_{c}^{256}$, and other models are fused RGB-D models. All fused models outperform the original RGB model. The performance of the best model, ${\cal N}_{conv(c_{256},d_{256})}^{256}$, is shown in bold.}
    \label{table:fop}
    \centering
    \scriptsize
    \begin{tabular} {@{}lccccc@{}}
    \toprule
    Models ~~~~ & {$\color{blue} {\cal N}_{c}^{256}$} ~~~~~& ${\cal N}_{conv(c_{256},d_{256})}^{256}$ & ${\cal N}_{sum(c_{256},d_{256})}^{256}$ & ${\cal N}_{MLP(c_{256},d_{256})}^{256}$ & ${\cal N}_{ca(c_{256},d_{256})}^{256}$\\
    \midrule
    AP  &  {\color{blue} 56.1}  ~~~~~ & \textbf{57.7} & 57.4 & 57.0 & 56.8  \\
    \bottomrule
    \end{tabular}
\end{table}

\subsection{Fusion Approaches' Performance}\label{insight}
The influence of various fusion approaches, presented in Section~\ref{SS:FF}, on the object detection performance is evaluated in this subsection. First, the fusion that keeps the fused feature encoder size the same as for the original model, ${\cal N}_c^{256}$, is considered. Table~\ref{table:fop} compares the performance of this fusion approach. All fused models outperform the original RGB unimodal network,  ${\cal N}_c^{256}$. The convolution model, ${\cal N}_{conv(c_{256},d_{256})}^{256}$, provides the best performance with improvement of $1.6\%$ in AP of $57.7\%$ compared with $56.1\%$ of the original RGB model, ${\cal N}_c^{256}$. This improvement can be explained by the spatial information used during dimensionality reduction in $conv$, as opposed to $MLP$, where RGB-D feature channels are only blended per pixel. The experiments with a deeper $conv$ fusion degrade object detection performance, especially for small objects. These observations support the importance of the receptive field size of the $conv$ fusion network. Note that the simple $ca$ fusion provides the least significant performance improvement between all the tested fused models. 

Table~\ref{table:fop-2} compares fusion operations without accounting for network capacity. Notice that the concatenation fusion outperforms other tested approaches at the expense of increased model capacity. 
Notice that both $MLP$ and $conv$ fusion operations' input is a concatenated RGB-D feature vector, $\oplus(x_c,x_d)$, with dimensionality $512$. However, further processing reduces the feature dimensionality to $256$. This dimensionality reduction results in information loss, which might be significant for object detection. All fused network architectures outperform original RGB unimodal, ${\cal N}_c^{256}$, and the ${\cal N}_{\oplus(c_{256}, d_{256})}^{512}$ performs the best as it uses all $512$ features without dimensionalty reduction.

\begin{table}[tb]
    \caption{The object detection performance of evaluated mid-level fusion approaches using AP metric without accounting for model capacity. Note that feature dimensionality reduction from $512$ to $256$ feature vector leads to degradation of the fused models' performance. This indicates that considering the availability of sufficiently large computational resources, the straightforward concatenation of RGB-D features could be the best fusion approach.}
    \label{table:fop-2}
    \centering
    \scriptsize
    \begin{tabular} {@{}lccccc@{}}
    \toprule
    Feature Fusion ~~~~ & ${\cal N}_{\oplus(c_{256},d_{256})}^{512}$ & ${\cal N}_{conv(c_{256},d_{256})}^{256}$ & ${\cal N}_{sum(c_{256},d_{256})}^{256}$ & ${\cal N}_{MLP(c_{256},d_{256})}^{256}$ & ${\cal N}_{ca(c_{256},d_{256})}^{256}$\\
    \midrule
    AP  & \textbf{58.7} & 57.7 & 57.4 & 57.0 & 56.8  \\
    \bottomrule
    \end{tabular}
\end{table}

\subsection{The Best Fused Concatenated Model Performance}\label{opl}
The experiments' results, summarised in Table~\ref{tab:object_detection_performance} show that ${\cal N}_{\oplus(c_{256},d_{256})}^{512}$  in $4^{th}$ line with features of dimensionality $512$ is the best fused model. For rigorous statistical analysis, this model is compared with the RGB unimodal, ${\cal N}_c^{512}$, in the $3^{rd}$ line of Table~\ref{tab:object_detection_performance}. The  fused model, ${\cal N}_{\oplus(c_{256},d_{256})}^{512}$, improved the AP of the RGB model, ${\cal N}_c^{512}$ in the $3^{rd}$ from $56.9\%$ to $58.7\%$.
Furthermore, comparing the performance of the best-fused model, ${\cal N}_{\oplus(c_{256},d_{256})}^{512}$, with ${\cal N}_c^{512}$ and ${\cal N}_c^{256}$, notice that the performance improvement is not associated with the increased network capacity but with the RGB-D feature fusion.
The performance of the depth-only model, $N_{d}^{256}$, is considerably lower than the RGB model, $N_{c}^{256}$. However, depth information enables auxiliary features not presented in RGB, and therefore,  their fusion improves object detection performance for all tested object categories. 

\begin{figure}[!tb]
    \centering
    \includegraphics[width=10cm]{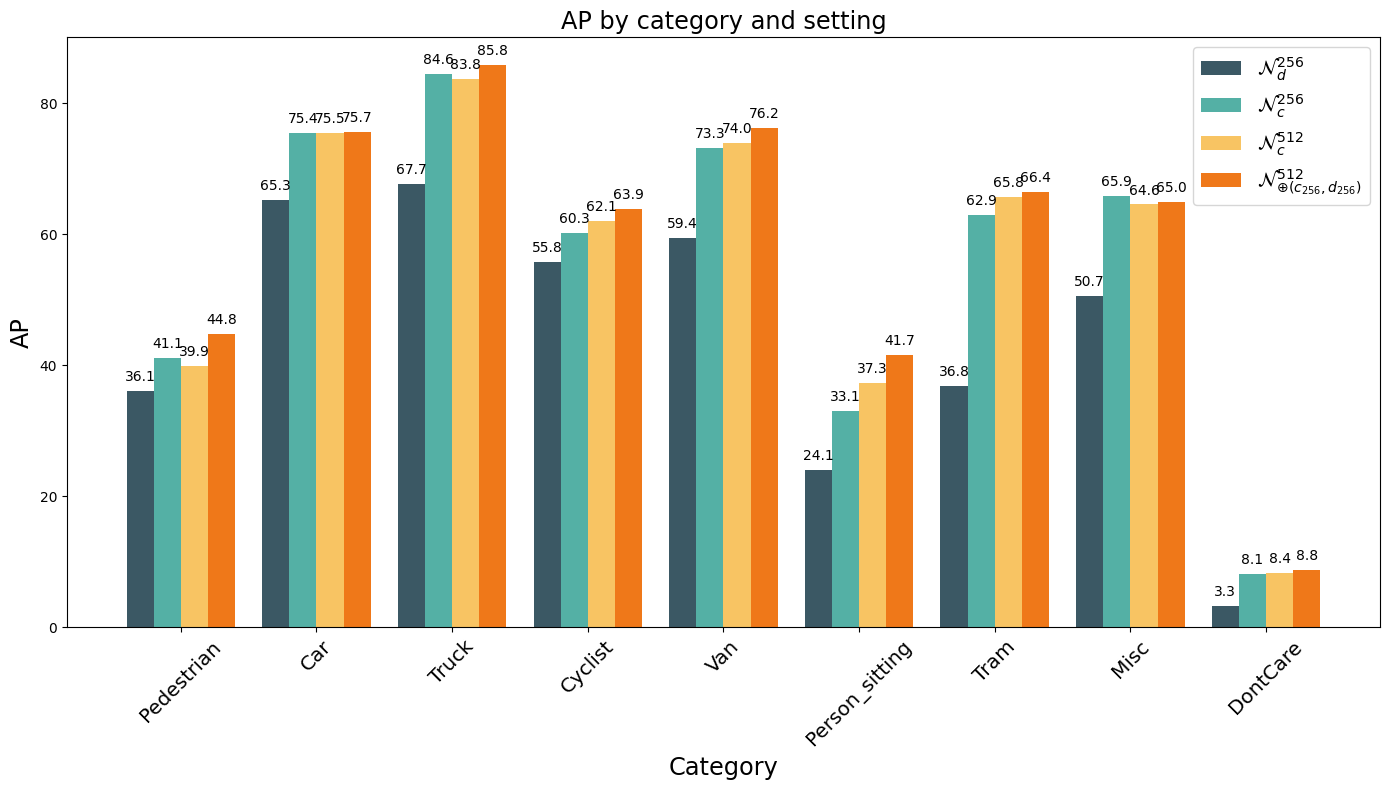}
    \caption{Visualization of the object detection performance of the unimodal and best fusion models per object category from Table~\ref{table:main_per_obj}. Notice that the models' average performance are as follows: ${\cal N}_d^{256} < {\cal N}_c^{256} < {\cal N}_c^{512} < {\cal N}_{\oplus(c_{256}, d_{256})}^{512}$. }
    \label{fig4}
\end{figure}

Table~\ref{table:fop} and Fig.~\ref{fig4} show that the fused model, ${\cal N}_{\oplus(c_{256},d_{256})}^{512}$, improves the AP performance for all considered object category. Notice that the object detection performance is improved for both small and large object categories. For example, the AP for the ``Person Sitting'' object category is improved from $37.3\%$ to $41.7\%$, and for the ``Van'' object category, from $74.0\%$ to $76.2\%$ compared with the model ${\cal N}_c^{512}$. A similar comparison with the original model ${\cal N}_c^{256}$ is even more prominent: the AP for the ``Person Sitting'' object category is improved from $33.1\%$ to $41.7\%$, and for the ``Van'' object category, from $73.3\%$ to $76.2\%$ and for pedestrians from $41.1\%$ to $44.8\%$.

\begin{table}[tb]
    \caption{The object detection performance of models using AP metric for all tested object categories. The same information is visually depicted in Fig.~\ref{fig4}. Note that the best fusion architecture ${\cal N}_{{\oplus}(c_{256},d_{256})}^{512}$ achieves statistically significant object detection performance improvement of $3.7\%$ for small objects such as "Pedestrian", and improvement of $8.6\%$ for "Person Sitting" category. The improvement of $2.9\%$ is achieved for large objects such as "Van". The best performance is marked in bold.}
    \label{table:main_per_obj}
    \centering
    \begin{tabular} {@{}lccccccccc@{}}
    \toprule
    Method/AP ~~~~ & Person & Pedestrian & Tram & Car & Truck & Cyclist & Van & Misc & Dont \\
    & Sitting & & & & & & & & Care \\
    \midrule
    ${\cal N}_d^{256}$ & 24.1 & 36.1 & 36.8 & 65.3 & 67.7 & 55.8 & 59.4 & 50.7 & 3.3 \\
    ${\cal N}_c^{256}$ & 33.1 & 41.1 & 62.9 & 75.4 & 84.6 & 60.3 & 73.3 & \textbf{65.9} & 8.2 \\
    ${\cal N}_c^{512}$ & 37.3 & 39.9 & 65.8 & 75.5 & 83.8 & 62.1 & 74.0 & 64.6 & 8.4 \\
    ${\cal N}_{\oplus(c_{256}, d_{256})}^{512}$  & \textbf{41.7} & \textbf{44.8} & \textbf{66.4} & \textbf{75.6} & \textbf{85.8} & \textbf{63.9} & \textbf{76.2} & 65.1 & \textbf{8.7} \\
    \bottomrule
    \end{tabular}
\end{table}

\section{Summary}\label{Concl}
This work presents a comprehensive study on enhancing automotive target detection via diffusion models using the fused RGB-D data. The advantages of fusing RGB and depth information over unimodal data are demonstrated. The proposed fusion approach uses a low-complexity concatenation feature fusion and achieves superior detection performance of various object categories.
The ablation studies demonstrate the efficiency of the proposed multimodal fusion strategy. It is shown that the improved object detection performance is associated with the efficient use of complementary information from RGB and depth sensors and not increased data dimensionality.
This result demonstrates the potential of the proposed FusedDiffusionDet network to improve autonomous vehicle sensing capabilities by enabling more accurate and reliable detection in complex driving environments.
The object detection performance of the proposed FusedDiffusionDet network is validated using the KITTI dataset, and the detection performance improvement of $3.7\%$ for small objects, such as pedestrians, and $2.9\%$ for large objects, such as vans, was demonstrated. 

To conclude, (i) mid-level fusion between RGB and depth provides significant object detection performance improvement, (ii) fusion performed as a simple concatenation between RGB and depth features provides the best object detection performance using the KITTI dataset and (iii) low-level fusion using ResNet backbone is inefficient.

\bibliographystyle{splncs04}
\bibliography{main}
\end{document}